\documentclass[10pt,twocolumn,letterpaper]{article}

\usepackage[pagenumbers]{cvpr} % To force page numbers, e.g. for an arXiv version

\usepackage{graphicx}
\usepackage{amsmath}
\usepackage{amssymb}
\usepackage{booktabs}
\usepackage{gensymb}  % degree
\usepackage{bm}  
\usepackage{enumitem}
\usepackage{microtype}

\usepackage[pagebackref,breaklinks,colorlinks]{hyperref}

\iftoggle{cvprfinal}{%
    \def\github{\url{https://github.com/jnyborg/tpe}}
}{%
    \def\github{\url{https://anonymous.4open.science/r/tpe-C5F3}}
}
\iftoggle{cvprfinal}{%
    \def\acknowledgements{Joachim Nyborg is funded by the \emph{Innovation Fund Denmark} under reference \emph{8053-00240}.
    We acknowledge the E-OBS dataset from the EU-FP6 project UERRA (\url{https://www.uerra.eu}) and the Copernicus Climate Change Service, and the data providers in the ECA\&D project (\url{https://www.ecad.eu}).}
}{%
    \def\acknowledgements{We acknowledge the E-OBS dataset from the EU-FP6 project UERRA (\url{https://www.uerra.eu}) and the Copernicus Climate Change Service, and the data providers in the ECA\&D project (\url{https://www.ecad.eu}).}
}

\usepackage[capitalize]{cleveref}
\crefname{section}{Sec.}{Secs.}
\Crefname{section}{Section}{Sections}
\Crefname{table}{Table}{Tables}
\crefname{table}{Tab.}{Tabs.}

\def\cvprPaperID{34} % *** Enter the EarthVision Paper ID here
\def\confName{EarthVision}
\def\confYear{2022}

\begin{document}

\title{Generalized Classification of Satellite Image Time Series\\ with Thermal Positional Encoding}

\author{Joachim Nyborg$^{1,3}$ \quad Charlotte Pelletier$^{2}$ \quad Ira Assent$^{1}$  \vspace{.2cm} \\ 
$^{1}$  Department of Computer Science, Aarhus University, Denmark\\
$^{2}$  IRISA UMR 6074, Univ. Bretagne Sud, France\\
$^{3}$  FieldSense A/S, Denmark\\
{\tt\small jnyborg@cs.au.dk, charlotte.pelletier@univ-ubs.fr, ira@cs.au.dk}
}

\maketitle

%%%%%%%%% ABSTRACT
\begin{abstract}
Large-scale crop type classification is a task at the core of remote sensing efforts with applications of both economic and ecological importance. Current state-of-the-art deep learning methods are based on self-attention and use satellite image time series (SITS) to discriminate crop types based on their unique growth patterns. However, existing methods generalize poorly to regions not seen during training mainly due to not being robust to temporal shifts of the growing season caused by variations in climate. To this end, we propose Thermal Positional Encoding (TPE) for attention-based crop classifiers. Unlike previous positional encoding based on calendar time (\eg day-of-year), TPE is based on thermal time, which is obtained by accumulating daily average temperatures over the growing season. Since crop growth is directly related to thermal time, but not calendar time, TPE addresses the temporal shifts between different regions to improve generalization. We propose multiple TPE strategies, including learnable methods, to further improve results compared to the common fixed positional encodings. We demonstrate our approach on a crop classification task across four different European regions, where we obtain state-of-the-art generalization results. Our source code is available at \github{}.
\end{abstract}

%%%%%%%%% BODY TEXT
\section{Introduction}
\label{sec:intro}
The increase in openly accessible satellite image time series (SITS) has led to the development of deep learning models using remote sensing data that has significantly improved the state of the art in SITS classification tasks.
Among these, crop type classification has numerous applications of economic and ecological importance, such as environmental monitoring, food security, and crop price prediction. Time series data is particularly valuable for crop classification, as it enables models to capture crop \textit{phenology}, \ie the progression of growth over time which characterizes different crop types.
Specialized deep learning models for the task thus focus on the temporal aspect of the problem, proposing models based on neural network components that process time, such as temporal convolutions~\cite{pelletier2019temporal, zhong2019deep}, recurrent layers~\cite{ndikumana2018deep, russwurm2017temporal, ienco2017land, minh2018deep}, or most recently self-attention~\cite{russwurm2020self, garnot2020satellite, garnot2020lightweight}.

\begin{figure}[t]
\centering
\includegraphics[width=1.0\linewidth]{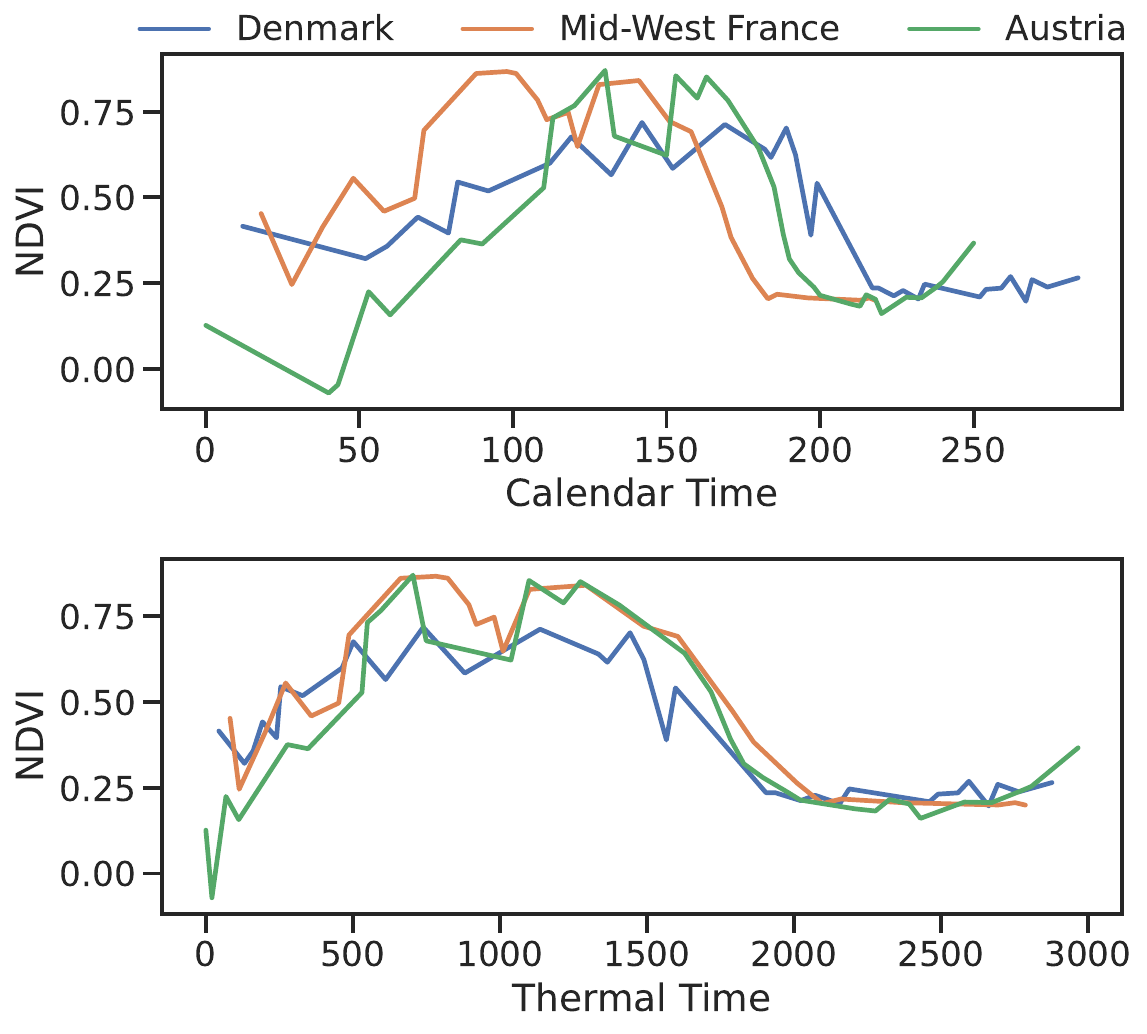}
\caption{Winter wheat NDVI in different European regions with calendar time and thermal time. 
With thermal time, temporal shifts of crop growth in different regions are greatly reduced.}
\label{fig:ndvi}
\end{figure}
 
Since the growth patterns of crops are similar in different regions of the world~\cite{hao2020transfer}, 
it is reasonable to expect that models trained in one region can generalize to another. However, recent works have found that existing models generalize poorly to other regions than those seen during training~\cite{nyborg2021timematch, lucas2021bayesian}. Part of the challenge in generalization is the variability in climate which causes different timing of crop growth~\cite{franch2015improving}. For example, in cooler regions, crops reach their growth stages later than in warmer regions, which models must account for to generalize~\cite{nyborg2021timematch}.

To model the progression of time, the predominant approach in existing models is to use calendar time to include temporal context, either during pre-processing to interpolate the data into regular temporal sampling~\cite{pelletier2019temporal, russwurm2020self, interdonato2019duplo, wang2021phenology} or as an explicit additional input~\cite{russwurm2018multi, rustowicz2019semantic}.
Notably, state-of-the-art methods based on self-attention input calendar time via \textit{positional encoding}~\cite{garnot2020satellite, garnot2021panoptic}. Since self-attention is position agnostic~\cite{vaswani2017attention}, 
this provides explicit positional information about the temporal location of images within the growing season. 
This helps crop classification as the particular timing for the phenological events of a crop type can be an important clue in its classification, \eg to distinguish spring wheat from winter wheat.
However, the phenological calendar timing of one region is not generally shared with other regions due to temporal shifts, which causes existing models to generalize poorly~\cite{kerner2020resilient, nyborg2021timematch}.

To overcome this challenge, we propose Thermal Positional Encoding (TPE) to improve the generalization of crop classifiers.
Our core idea is to use a representation that captures the climatic variation affecting growth rates without relying on calendar time.
To this end, we propose positional encoding based on \textit{thermal time}~\cite{mcmaster1997gdd, mederski1973accumulated} for self-attention models.
Thermal time is typically measured for crops by units of \textit{Growing Degree Days} (GDD)~\cite{mcmaster1997gdd, miller2001using, yang1995mathematical, hao2020transfer}, computed by accumulating daily average temperatures above a baseline.  
As crop growth is directly related to the accumulation of heat over the growing season~\cite{franch2015improving, jones2013plants}, an earlier crop growth corresponds to an earlier increase in GDD and vice versa.
This is illustrated in Figure~\ref{fig:ndvi} using normalized difference vegetation index (NDVI) to display winter wheat phenology in three different regions. 
Thermal time improves generalization of models by making SITS from different regions invariant to temporal shifts. At the same time, it provides a temporal location of images which allows thermal time to directly replace calendar time in crop classifiers.
%At the same time, since thermal time is computed by daily accumulation, models can still capture the class-wise phenological timings as with calendar time.

%
%At the same time, models can still capture the class-wise phenological timings 
%
%
%At the same time, thermal time still captures the progression of time via the daily accumulation, 
%which enables models to capture the class-wise phenological timings which was also possible with calendar time.

%via the daily accumulation, and models can still capture the class-specific phenological timings.
%while still capturing crop-specific phenological timings.

To encode positional information, existing works generally use sinusoidal encoding~\cite{vaswani2017attention}.
However, as this approach is predefined and not learned, it lacks flexibility and may not capture crop-specific positional information.
In this paper, we propose multiple TPE methods to encode thermal time in a data-driven way.
By learning an encoding function instead of, \eg an embedding vector for each position~\cite{devlin2018bert, radford2018gpt1}, our approach is inductive. This allows us to handle when the thermal time of test regions differs from that of training, which is common in practice.
We evaluate our approach on a crop classification task across four different European regions on the TimeMatch dataset~\cite{nyborg2021timematchdataset}, containing Sentinel-2 SITS expanded with daily temperature data, and demonstrate that we obtain state-of-the-art generalization results in new regions. Our main contributions are:
\begin{itemize}
    \item We propose the use of thermal time in crop classification to increase robustness to temporal shifts and improve generalization.
    \item We propose TPE methods, which are based on thermal time and can easily be implemented in recent attention-based crop classifiers.
    \item We demonstrate that TPE greatly improves generalization across four different European regions.
\end{itemize}
%We evaluate our method for parcel classification on the large-scale TimeMatch dataset~\cite{nyborg2021timematch}, containing Sentinel-2 SITS from four European countries, and demonstrate that TPE greatly improves generalization in new regions. In summary:

\section{Related Work}
\label{sec:related}

\paragraph{Satellite Image Time Series Classification.}
Multiple traditional machine learning approaches, such as random forests or support vector machines, have been applied to crop classification~\cite{vuolo2018much, wang2019croptype, wardlow2008large, inglada2015assessment}. 
These approaches require input features to be extracted by hand. For instance, a widely used feature is NDVI, combining the red and near-infrared spectral bands, which relates to the photosynthesis of crops~\cite{tucker1979red}.
Other works also include phenological features~\cite{jia2014land, valero2016production} or meteorological information~\cite{zhong2014efficient}. Although these handcrafted features are robust and interpretable, deep learning approaches are mostly employed as they enable the automatic extraction of richer features from raw SITS.
Deep convolutional networks have been widely applied to process the spatial dimensions of the data~\cite{kussul2017deep, russwurm2018multi}, while the temporal dimension has been processed by recurrent units~\cite{russwurm2017temporal, ndikumana2018deep}, 1D convolutions~\cite{zhong2019deep, pelletier2019temporal}, or combinations thereof~\cite{interdonato2019duplo, russwurm2018multi}.
Recently, self-attention~\cite{vaswani2017attention} has led to significant improvements in
pixel~\cite{russwurm2020self} and parcel classification~\cite{garnot2020satellite, garnot2020lightweight},
as well as semantic and panoptic segmentation~\cite{garnot2021panoptic}. 
Since self-attention is position-agnostic, existing works use sinusoidal positional encoding~\cite{vaswani2017attention} of calendar time to capture the position of images in the growing season. We propose positional encoding based on thermal time~\cite{mederski1973accumulated, mcmaster1997gdd} to improve the generalization of the promising self-attention mechanism.

\paragraph{Domain Generalization for SITS.}
Several prior works have reported that existing crop classification models fail to generalize across space and time due to not being robust to temporal shifts of the growing season~\cite{lucas2020unsupervised, nyborg2021timematch, kerner2020resilient, wang2021phenology}. 
This problem has mainly been tackled by unsupervised domain adaptation (UDA), where models are trained with labeled data from a source region and unlabeled data from a target region~\cite{tuia2016domain}. Phenology Alignment Network~\cite{wang2021phenology} addresses this problem by learning domain-invariant features obtained with a maximum mean discrepancy loss~\cite{tzeng2014deep} for the unlabeled target data. TimeMatch~\cite{nyborg2021timematch} obtains further improvements by directly estimating the temporal shift of the target region, and utilizing the shift estimation to train with pseudo-labels for the unlabeled target region.
Our setting differs from UDA, as we do not aim to adapt models to particular regions by training with unlabeled data, but to improve the generalization of a crop classifier model trained with labeled data from multiple areas to any new region.

Most similar to our work, Kerner~\etal~\cite{kerner2020resilient} improve the generalization of crop classifiers by inputting 
satellite data at specific time steps which correspond to particular growth stages (greenup, peak, and senescence), computed from the NDVI sequence for each input. By dynamically selecting these time steps, this approach can account for temporal shifts of the growing season, but information is lost since the complete time series is not involved in the prediction.
In comparison, we aim to train self-attention models which attend to the most relevant time steps in the complete time series automatically by incorporating thermal time.

\paragraph{Positional Encodings.}
A vast literature exists in positional encoding for the self-attention mechanism. 
Absolute positional encoding is most widely used. In the original Transformers~\cite{vaswani2017attention}, vectors are encoded from the absolute position in the sequence by sinusoidal functions, but this approach is less flexible as the vectors are fixed and not learned. To overcome this issue, a common approach is to learn an embedding vector for each position~\cite{devlin2018bert, radford2018gpt1} similar to word embeddings, but this approach requires all possible positions to be seen during training to ensure all the embeddings are updated by gradient descent.
This is ill-suited for irregularly sampled SITS, which does not guarantee that all possible (calendar or thermal) positions are available for training.
Instead, approaches that learn a function that maps positions to vectors~\cite{neishi2019relation, liu2020learning, li2021learnable} do not have this requirement and can thus generalize to unseen positions at test time. We therefore build upon these in this paper.

Another line of work is relative positional encoding~\cite{shaw2018self, huang2018improved, dai2019transformer}, which encodes the positional difference between each pair in the input sequence instead of the absolute position of individual elements.
While relative positions can be more relevant than absolute in other tasks, in SITS classification, the absolute position is crucial information. For example, a satellite image taken during the winter will not contain the same information about crop growth compared to an image from the spring, which cannot be captured by only the relative positions, \eg the difference in days between the two images. 
Thus, we focus on absolute positional encoding in this work.

\section{Self-Attention for Crop Classification}
\label{sec:method}
In crop classification, we are given a satellite image time series $\bm x = [\bm x^{(1)}, \dots, \bm x^{(T)}]$, where $T$ is the length of the time series.
The goal of the classification task is to associate $\bm x$ with one of $K$ classes. In our setting, each $\bm x^{(t)} \in \mathbb{R}^{T \times N \times C}$ consists of a sequence of $N$ pixels of $C$ spectral bands within a \textit{parcel}, \ie, a homogeneous agricultural plot of land.
This approach requires parcel shapes to be available in the region for classification, which is widely available in the European Union (EU)~\cite{schneider2021eurocrops} or can alternatively be acquired by a segmentation step~\cite{stoian2019land, garnot2021panoptic}.

Our goal is to improve the generalization of existing crop classifiers by accounting for temporal shifts of the growing season.  Owing to its state-of-the-art performance, we build upon the PSE+LTAE model~\cite{garnot2020lightweight}. 
The network consists of the Pixel-Set Encoder (PSE) and the Lightweight Temporal Attention Encoder (LTAE).
Given a randomly sampled pixel-set of size $S$ among the $N$ available pixels of an input $\bm x$, the PSE handles the spatial and spectral context of SITS by processing each time step individually to a sequence of embedding vectors  $\bm e = [\bm{e}^{(1)}, \dotsc, \bm{e}^{(T)}] \in \mathbb{R}^{T \times D}$, where $D$ is the embedding dimension. 
PSE does not process the temporal dimension.  We thus focus on handling temporal shifts in the LTAE module. Given $\bm{e}$, LTAE extracts temporal features using a simplified version of the multi-headed self-attention, as we describe next.

\paragraph{Self-Attention.}
In the original Transformer model~\cite{vaswani2017attention}, self-attention is computed with a query-key-value triplet $(\bm{q}^{(t)}, \bm{k}^{(t)}, \bm{v}^{(t)})$ for each element in the input sequence using three fully-connected layers. 
The output is a sequence where each element is a sum of all values $\bm{v}^{(t)}$ weighted by their attention score. The attention scores for a time step $t$ are computed as the similarity (dot product) between all keys and the query $\bm{q}^{(t)}$, re-scaled by a softmax layer.
The computation of the query-key-value triplets can be performed in parallel, which enables the Transformer model to take full advantage of GPUs for a significant speed increase compared to the sequential computation of recurrent neural networks (RNN). 
In multi-headed self-attention, the triplets are computed multiple times in parallel with different parameters, or ``heads'', which further increase efficiency and also the representational capacity as each head can specialize in different parts of the sequence.

\paragraph{Sinusoidal Positional Encoding.}
As the self-attention mechanism is position-agnostic~\cite{vaswani2017attention}, various positional encodings (PE) have been introduced to capture positional information. This is typically done by mapping scalar positions to a vector, either by learning or by heuristics, and adding each embedding vector with their positional encoding $\bm{e}^{(t)} + \bm{p}^{(t)}$ before applying self-attention. The original Transformer model~\cite{vaswani2017attention} uses a fixed sinusoidal encoding with predefined wavelengths, defined as:
\begin{equation}
    \label{eq:sinusoidal}
    \bm{p}^{(t)} = \left[\sin(\omega_i t), \cos(\omega_i t)\right]_{i=1}^{D/2} 
\end{equation}
where $\omega_i = (1/\tau)^{2i/D}$ and $\tau = 10000$.

\paragraph{Lightweight Temporal Attention Encoder.}  
While the original self-attention maps the input embeddings $\bm e$ to an output sequence of embeddings, the goal of SITS classification is to map the entire time series into a single embedding.
To address this, the LTAE module~\cite{garnot2020lightweight} modifies the self-attention mechanisim by replacing the queries $\bm{q}^{(t)}$ with a single learnable ``master'' query $\bm{\hat{q}}$, resulting in a single output embedding instead of a sequence.
The computation is also made more lightweight by employing a channel grouping strategy~\cite{wu2018group}, where each attention head operates on its own subset of input channels. 
The LTAE module uses the sinusoidal PE (Equation~\ref{eq:sinusoidal} with $\tau=1000$) but encodes the 
day of the year $\mathrm{day}^{(t)}$ instead of the position index $t$. This enables the model to account for the inconsistent temporal sampling of SITS, but also introduces problems with handling temporal shift~\cite{nyborg2021timematch}.

\begin{figure*}[ht]
\centering
\includegraphics[width=0.9\linewidth]{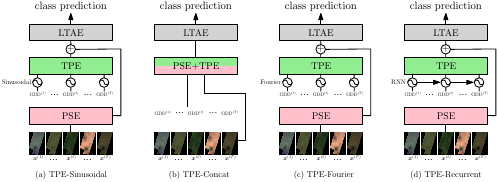}
\caption{Schematic illustration of our Thermal Positional Encoding (TPE) methods with the PSE+LTAE model~\cite{garnot2020lightweight}.}
\label{fig:method}
\end{figure*}
\section{Method}    
\label{sec:thermaltime}
In this work, we observe that the positional encoding used by the LTAE module has two issues. First, since it encodes calendar time, it introduces the temporal shift problem as displayed in Figure~\ref{fig:ndvi}. While calendar time is useful to identify the crop types in a particular region, it hinders generalization to new regions~\cite{kerner2020resilient}. For example, while spring and winter crops can be similar in appearance, they are easily separated by the timing of their growth stages as spring crops are planted later in a growing season than winter crops. However, because of temporal shifts, the same time positions of spring crops could represent winter crops in another region. Without any way of accounting for temporal shifts, calendar time positional encodings are unlikely to generalize.
Second, since the positional encoding is fixed and not learned, it prevents the model from taking advantage of end-to-end training the encoding function to further improve generalization~\cite{liu2020learning, li2021learnable, radford2018gpt1}. 

A possible remedy to the first issue is to augment the training data with random temporal shifts, such as ShiftAug, a SITS augmentation technique proposed in~\cite{nyborg2021timematch}, so that the model does not learn to associate a specific position with the phenological events seen in the training data. While this solution increases the invariance of the model to temporal shifts, the temporal shift is in some cases an important clue to distinguish crop types---such as the spring and winter crops. Instead, we want models that are shift-invariant between different regions, but shift-variant within the same region. That is, we want models which can use class-wise temporal shifts for classification but are unaffected by temporal shifts of the growing season.

To address the second issue, a common alternative to the fixed sinusoidal encodings is to treat each position as a discrete token that can be uniquely represented as a learnable vector~\cite{devlin2018bert, radford2018gpt1, gehring2017convolutional}. While this approach enables the model to learn the positional encoding from data, it fails to generalize to positions not encountered during training. This is an issue for high-resolution SITS, as we typically do not have an observation for every possible position. For example, the Sentinel-2 satellites acquire images every five days. Moreover, images with high cloud coverage are often filtered, further reducing the positions available. 
In comparison, sinusoidal positional encoding is more practical for SITS, as an encoding vector is well-defined for every position independent of the training data.

\subsection{Thermal Positional Encodings}
We argue that successful positional encoding for SITS should meet the following requirements:
\begin{enumerate}[label={(\arabic*)}]
    \item Making SITS from \textit{different} regions shift-\textit{invariant} to address the temporal shift problem.
    \item Making SITS from the \textit{same} region shift-\textit{variant} by providing absolute information of where an observation is located in the growing season.
    \item Must be inductive to be able to handle positions not seen during training.
    \item Being data-specific and thus learnable.
\end{enumerate}

While the LTAE sinusoidal positional encoding based on calendar time meets the second and third requirements, it is not invariant to temporal shifts between different regions or trainable which violates the first and fourth requirements. To address this, we replace calendar time with thermal time to meet both the first and second requirements and propose four TPE strategies, including learnable methods to meet the third and fourth requirements.

%. Some of which learn the positional encoding function for thermal time and meet the third and fourth requirements.

\paragraph{Thermal time.} When studying crop phenology, thermal time is a good proxy for the rate of crop growth~\cite{mcmaster1997gdd, franch2015improving, trudgill2005thermal}. Thermal time is typically measured in units of \textit{growing degree days} (GDD). 
The GDD measured at a time $t$ is computed by accumulating daily average temperatures above a baseline: 
\begin{align}
    \label{eq:gdd}
    \mathrm{GDD}^{(t)} = \sum_{i=1}^{t} \max\left( \frac{T_{min}^{(i)} + T_{max}^{(i)}}{2} - T_{base}, 0 \right)
\end{align}
where $T_{min}^{(i)}, T_{max}^{(i)}$ is the minimum and maximum temperatures for day $i$, accumulated for all the previous days $i=1,2,\dots,t$. Temperature values are often clipped to a range $[T_{base}, T_{cap}]$ chosen depending on the crop type.
Since we do not know the crop type of the input beforehand, we choose standard values
$T_{base}=0$ and $T_{cap}=30$~\cite{mcmaster1997gdd, miller2001using} for all crops, since growth typically stagnates below 0\degree C and does not grow any faster above 30\degree C.
We accumulate from the starting day of the input SITS, in our case January 1. Since GDD is computed by a cumulative sum, it is a monotonically increasing function and thus preserves the order of the input time series. This enables GDD to directly replace day of year for the time positions in the self-attention computation.
By replacing calendar time with thermal time, we can reduce the temporal shift of SITS between different regions while retaining the shift between classes within the same region and thereby satisfy the first and second requirements. 

\paragraph{TPE Methods.}
We propose the following TPE methods to input thermal time to PSE+LTAE~\cite{garnot2020lightweight}. 
\begin{itemize}
    \item TPE-Sinusoidal: We replace calendar time with thermal time in the sinusoidal PE, but the encoding is not learned.
    \item TPE-Concat: We learn SITS and positional input embeddings jointly by concatenating thermal time to an intermediate feature of the PSE module.
    \item TPE-Fourier: We learn the sinusoidal PE function by the method proposed in~\cite{li2021learnable}.
    \item TPE-Recurrent: We learn a positional encoding function that captures the development in GDD by a recurrent neural network (RNN).
\end{itemize}
An overview of the TPE methods is shown in Figure~\ref{fig:method}.

\subsection{TPE-Sinusoidal}
To use GDD with the sinusoidal PE, we follow Equation~\ref{eq:sinusoidal} but replace $t$ with $\mathrm{GDD}^{(t)}$. 
The benefit of using the sinusoidal positional encoding for GDD is that an encoding vector is well-defined for every possible GDD value. 
This ensures that even if we train with only a subset of possible accumulated temperatures, a positional encoding exists 
for unseen positions at test time.
However, as the sinusoidal PE is fixed and not learned, it prevents the model from capturing data-specific positional information for the crop classification task.

%from discovering other positional encodings which could be useful for the crop classification task.

\subsection{TPE-Concat}
While the original Transformer network~\cite{vaswani2017attention} takes pre-trained word embeddings as inputs, in our case, the embeddings are learned by the PSE module, which is learned simultaneously to the LTAE module. Thus, we propose an alternative to positional encoding where 
the encoding for the SITS and positions are learned jointly by the PSE. 
In particular, for each time step $t$, we concatenate $\mathrm{GDD}^{(t)}$ to the intermediate PSE embedding $\bm{\hat{e}}^{(t)}$ before the final PSE output layer $\mathrm{MLP}_{2}$:
\begin{equation}
    \bm{e}^{(t)} = \mathrm{MLP}_{2}([\bm{\hat{e}}^{(t)} \mid \mid \mathrm{GDD}^{(t)}]),
\end{equation}
where $[\cdot\mid\mid\cdot]$ indicates concatenation. The PSE output layer $\mathrm{MLP}_2$~\cite{garnot2020satellite} 
is a multi-layer perceptron (MLP) consisting of a linear layer, batch normalization~\cite{ioffe2015batch}, and ReLU~\cite{nair2010rectified} activation function. We note that this approach is similar to the method of inputting extra parcel geometric features in the original PSE. By concatenating positions to the embedding function, TPE-Concat removes the need for complex positional encoding functions, which may be more beneficial for SITS.

\subsection{TPE-Fourier}
Li~\etal~\cite{li2021learnable} propose a learnable PE based on Fourier features~\cite{rahimi2007random},   
which can also be viewed as a generalization of the sinusoidal PE. For a position $t \in \mathbb{R}$, the Fourier PE is computed by:
\begin{equation}
    \bm{r}^{(t)} = \frac{1}{\sqrt{D}}[\cos(\bm{W}_r t) \mid\mid \sin(\bm{W}_r t)],
\end{equation}
where $\bm{W}_r \in \mathbb{R}^{D/2}$ is a trainable vector. To give the representation additional capacity, the encoding is passed through an MLP:
\begin{equation}
    \bm{p}^{(t)} = \mathrm{MLP}(\bm{r}^{(t)})\bm{W}_p
\end{equation}
where $\mathrm{MLP}$ consists of a linear layer with GeLU~\cite{hendrycks2016gaussian} activation function, and $\bm{W}_p$ are parameters for projecting the representation to the dimension of the input embeddings.
The TPE-Fourier reveals whether it is more beneficial to learn the sinusoidal PE compared to the fixed TPE-Sinusoidal.

\subsection{TPE-Recurrent}
Compared to natural language processing (NLP), where positions typically increase linearly with the sequence length, GDD increases non-linearly over the growing season (see Figure~\ref{fig:gdd}), as a result of the higher daily temperatures during the summer than the winter.
It may therefore be beneficial not to only encode independent GDD values, but also incorporate previous values to account for different rates of crop growth over the year.
To handle this, we propose to use an RNN to learn the positional encoding. %, as a recurrent unit naturally captures temporal development.
RNNs have been successfully used for positional encoding in NLP tasks~\cite{neishi2019relation, liu2020learning}. We follow the RNN approach of Liu~\etal~\cite{liu2020learning}. In particular, we use a GRU~\cite{cho2014learning}, which computes its output $\bm{h}^{(t)} \in \mathbb{R}^{H_{out}}$ for each time step $t$ given an input $\bm{z}^{(t)} \in \mathbb{R}^{H_{in}}$ and the previous hidden state $\bm{h}^{(t-1)}$ by:
% to encode GDD.
% Instead of learning to encode each GDD position independently as other TPE methods, it may also be beneficial to incorporate previous values to account for the changing rates of crop growth.
% To handle this, we propose to use an RNN to learn the positional encoding
% RNN have already been succesfully used for positional encoding in NLP tasks. We therefore follow the RNN approach of~\cite{liu2020learning}
% to encode GDD.
% In our work, we learn GDD embeddings to capture not only independent GDD positons, but also the history of values to account for the current rate of crop growth.

%, as a recurrent unit naturally captures temporal development in positions

%As in~\cite{liu2020learning, neishi2019relation}, we use an RNN model to learn the positional encoding. 
%Instead of learning to encode each GDD position independently, a recurrent unit naturally captures
%the change in GDD and thereby the growth rate, which may be beneficial to adapt to new regions.        

%As RNN, we use the GRU~\cite{cho2014learning}, which computes its output $\bm{h}^{(t)} \in \mathbb{R}^{H_{out}}$ for each time step $t$ given an input $\bm{z}^{(t)} \in \mathbb{R}^{H_{in}}$ and the previous hidden state $\bm{h}^{(t-1)}$ by:
\begin{equation}
    \bm{h}^{(t)} = \mathrm{GRU}(\bm{z}^{(t)}, \bm{h}^{(t-1)}).
\end{equation}
Then, we obtain a positional encoding with target dimension $D$ by a linear projection:
\begin{equation}
    \bm{p}^{(t)} = \bm{W}_p^{\top} \bm{h}^{(t)} + \bm{b}_p,
\end{equation}
where $\bm{W}_p \in \mathbb{R}^{H_{out} \times D}$ and $\bm{b}_p \in \mathbb{R}^{D}$. 
Instead of scalar values $\mathrm{GDD}^{(t)}$, we use vectorized positions as the inputs $\bm{z}^{(t)}$, which are obtained by obtained by the sinusoidal positional encoding of $\mathrm{GDD}^{(t)}$ (Equation~\ref{eq:sinusoidal}) as done in~\cite{liu2020learning}.
TPE-Recurrent learns a positional encoding that captures the temporal development in GDD, but is more computationally expensive due to the sequential computation of an RNN.

\section{Experiments}
\label{sec:experiments}

\subsection{Setup}

\begin{figure}[t]
\centering
\frame{\includegraphics[width=0.7\linewidth]{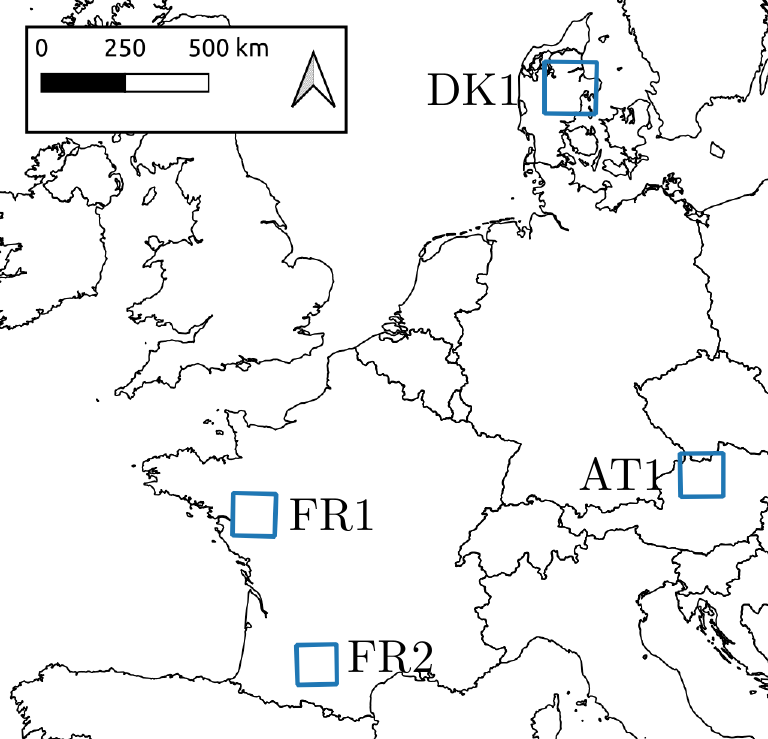}}
\caption{The geographical locations in Europe of the four Sentinel-2 tiles in the dataset~\cite{nyborg2021timematchdataset}. Figure adapted from~\cite{nyborg2021timematch}.}
\label{fig:tilemap}
\end{figure}

\begin{table*}[ht]
\begin{center}
\begin{tabular}{@{}l cc c cc c cc c cc c cc @{}}
\toprule 
& \multicolumn{2}{c}{AT1} && \multicolumn{2}{c}{DK1} && \multicolumn{2}{c}{FR1} && \multicolumn{2}{c}{FR2} && \multicolumn{2}{c}{Avg.} \\
Method & F1 & OA && F1 & OA && F1 & OA && F1 & OA && F1 & OA \\
\midrule
%\multicolumn{15}{l}{\textit{calendar time}} \\
PSE+LTAE~\cite{garnot2020lightweight} & 68.3 & 90.5 && 55.4 & 62.6 && 74.6 & 90.9 && 73.5 & 87.5 && 68.0 & 82.9 \\                                                    
%Don't concat PE & 76.0 & 89.5 && 53.8 & 63.0 && 74.9 & 89.2 && 80.0 & 89.8 && 71.2 & 82.9 \\
%GDD SPE don't concat & 84.9 & 94.4 && 81.5 & 86.7 && 82.6 & 91.9 && 81.2 & 89.8 && 82.6 & 90.7 \\
%GDD SPE+Conv & 85.4 & 94.8 && 78.5 & 85.7 && 81.2 & 91.2 && 81.6 & 90.1 && 81.7 & 90.5 \\                                                                                                                                                                                     
%GDD PE only key & 84.2 & 93.7 && 79.5 & 83.5 && 82.2 & 92.7 && 81.6 & 90.0 && 81.9 & 90.0 \\
+ w/o PE & 84.1 & 94.4 && 66.3 & 76.2 && 79.3 & 91.9 && 74.0 & 86.4 && 75.9 & 87.2 \\
+ w/ ShiftAug~\cite{nyborg2021timematch} & 84.2 & 94.1 && 71.6 & 78.5 && 83.9 & 93.3 && 79.8 & 89.4 && 79.9 & 88.8 \\
%\multicolumn{15}{l}{\textit{thermal time}} \\
\midrule
+ TPE-Sinusoidal & 85.6 & 94.7 && 78.7 & 84.8 && 83.0 & 92.6 && 81.1 & \textbf{90.4} && 82.1 & 90.6 \\
+ TPE-Concat & 85.7 & 94.7 && 78.6 & 83.1 && 85.1 & 93.3 && \textbf{81.4} & 89.6 && 82.7 & 90.2 \\
+ TPE-Fourier & 84.7 & 94.4 && 79.0 & \textbf{86.0} && 77.3 & 91.5 && 80.0 & 89.4 && 80.3 & 90.3 \\ 
+ TPE-Recurrent & \textbf{86.5} & \textbf{95.0} && \textbf{80.3} & 85.4 && \textbf{86.0} & \textbf{93.8} && 80.5 & 89.8 && \textbf{83.3} & \textbf{91.0} \\             
\midrule
%Upper bound & 94.7 & 97.5 && 92.6 & 93.9 && 93.2 & 96.5 && 87.5 & 94.0 && 92.0 & 95.5 \\  % PSE+LTAE
Upper-bound & 94.6 & 97.5 && 92.0 & 94.0 && 93.1 & 96.4 && 87.4 & 93.9 && 91.8 & 95.4 \\  % TPE-Recurrent

%+ TimeMatch GDD sinusoid & 85.7 & 94.7 && 84.1 & 89.6 && 84.4 & 93.5 && 81.3 & 90.2 && 83.9 & 92.0 \\
\bottomrule
\end{tabular}
\end{center}
\caption{Leave-one-region-out spatial generalization results in macro F1 score (F1) and overall accuracy (OA) (both in \%). Each column shows the classification results in a new region after training on the others.}
\label{tab:space_results}
\end{table*}
\paragraph{Dataset.}
We evaluate our approach on the TimeMatch dataset~\cite{nyborg2021timematchdataset} with Sentinel-2 L1C SITS from four different tiles: 33UVP (Austria), 32VNH (Denmark), 30TXT (mid-west France), and 31TCJ (southern France). We refer to these regions by AT1, DK1, FR1, and FR2, respectively. 
We display the locations of these tiles in Figure~\ref{fig:tilemap}.
The dataset contains all available observations of these tiles between January 1, 2017, and December 31, 2017, with cloud cover $\leq80\%$ and coverage $\geq 50\%$. 
The atmospheric bands (1, 9, and 10) are left out, keeping the remaining $10$ spectral bands. The 20m bands are bilinearly interpolated to 10m. 
\begin{figure}[h]
\centering
\includegraphics[width=0.8\linewidth]{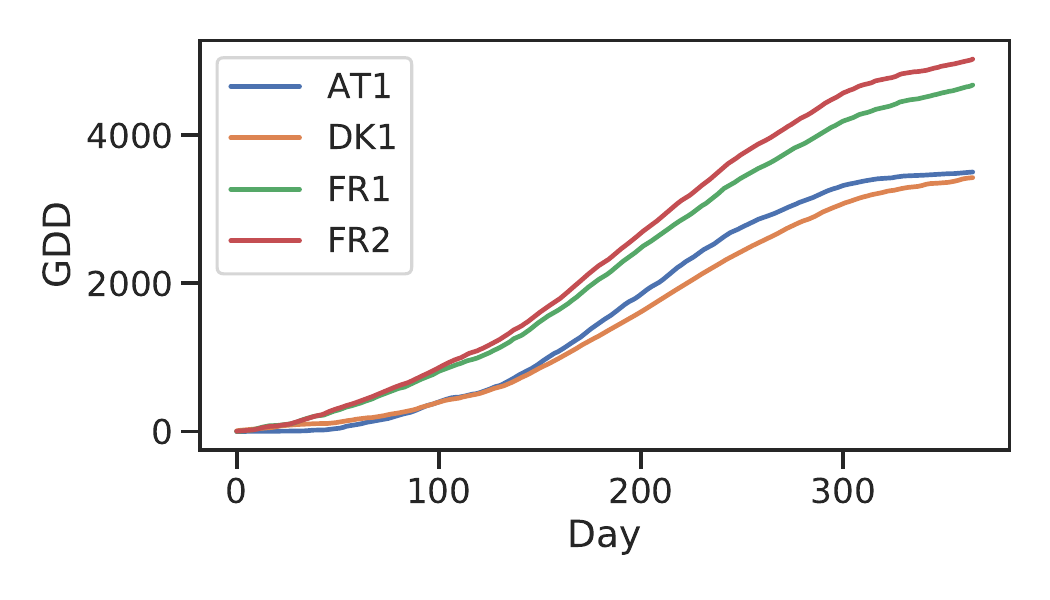}
\caption{The development of GDD on average in the different Sentinel-2 tiles from January 1 to December 31, 2017.}
\label{fig:gdd}
\end{figure}

The dataset is prepared for parcel classification by cutting the pixels of each parcel from the SITS using geo-referenced parcel shapes available from the Land Parcel Identification System (LPIS) in each country. The total amount of parcels is 280K with 15 crop classes.
The frequency of these classes varies greatly between tiles, for example, sunflowers are only frequent in the two France tiles. To ensure all tiles have enough samples of each class to learn their classification, we select the 9 crop types with at least 200 samples in all tiles: 
corn, horsebeans, meadow, spring barley, winter barley, winter rapeseed, winter triticale, winter wheat, and unknown. 
Here, the unknown class contains all parcels with crop type not of the other 8 classes.
Each tile has its own train/validation/test sets, created by assigning all parcels in a tile at random to these sets by a 70\%/10\%/20\% ratio.

We expand the TimeMatch dataset with weather information from the Europe-wide E-OBS dataset~\cite{cornes2018eobs}. 
We use the daily minimum and maximum temperature from the $0.1\degree$ regular grid of 2017 to compute GDD for each parcel, geo-referenced by the parcel centroid.
Figure~\ref{fig:gdd} displays the average GDD computed for the four regions, showing the southern France tile FR2 is the warmest and the Danish tile DK1 the coldest.

\paragraph{Implementation details.}
We follow the original implementation of PSE+LTAE~\cite{garnot2020lightweight}.
All models are trained for 100 epochs with a batch size of 128 on a single GTX 1080Ti GPU with Adam optimizer~\cite{kingma2014adam}. The learning rate is initialized to $1\mathrm{e}{-3}$ and 
decayed each epoch by cosine annealing~\cite{loshchilov2016sgdr}. We use weight decay of $1\mathrm{e}{-4}$. The 16-bit input pixels are normalized to $[0, 1]$ by dividing by $2^{16}-1$.
Our code is available at \github{}.

\paragraph{Experimental setup.}
To evaluate whether our proposed thermal positional encoding improves generalization to new regions, we adopt a leave-one-region-out setup where we hold one Sentinel-2 tile out for testing and train on the remaining. In contrast to the domain adaptation setup of TimeMatch~\cite{nyborg2021timematch}, where data is only available from one tile for training, our setup contains multiple different regions for training. In practice, we typically have many tiles available for training~\cite{schneider2021eurocrops}, so this setup allows us to evaluate against the naive approach of improving generalization by adding more training data.% and training a single deep network end-to-end.

\paragraph{Model variants.}
In comparison to TPE, we consider the following model variants:
\begin{itemize}
    \item \textit{PSE+LTAE}~\cite{garnot2020lightweight}. This is the baseline model which encodes calendar time (day of the year) with the sinusoidal positional encoding~\cite{vaswani2017attention}.
    \item \textit{w/o PE}. This is PSE+LTAE where self-attention is computed without any positional information.
    \item \textit{w/ ShiftAug}~\cite{nyborg2021timematch}. PSE+LTAE trained with calendar time augmented with random temporal shifts.% in $\pm 60$ days.
    \item \textit{Upper-bound}. We train the best performing TPE method (TPE-Recurrent) with all four available regions to obtain the results of a fully-supervised upper bound.
\end{itemize}

\subsection{Parcel Classification Results}

\begin{figure*}[ht]
    \centering
    \includegraphics[width=\linewidth]{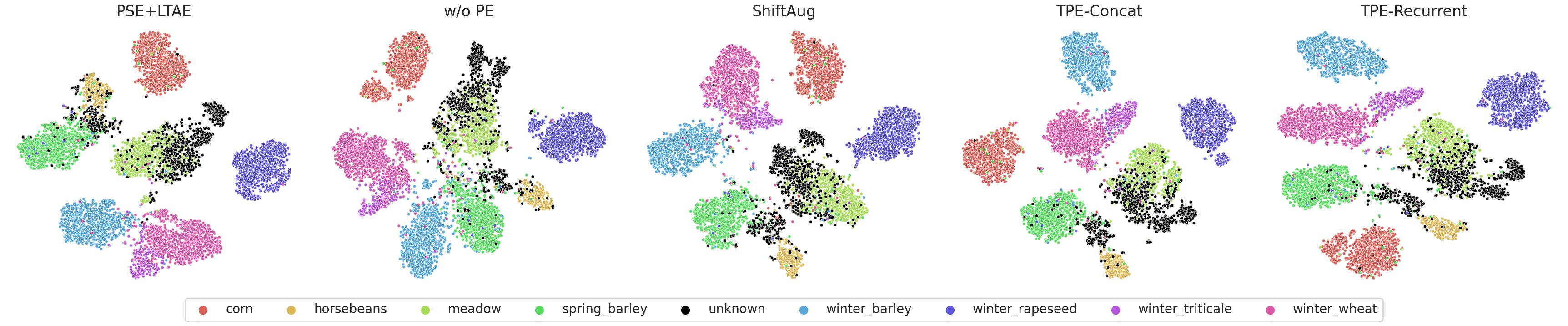}
    \caption{LTAE features of different methods embedded with t-SNE~\cite{van2008visualizing} for DK1 after training with the remaining regions.}
    \label{fig:tsne}
\end{figure*}
In Table~\ref{tab:space_results}, we detail the performance obtained for the leave-one-region-out spatial generalization experiments.
We report the class-averaged F1 score (F1) and the overall accuracy (OA). 
Compared to calendar time models (top), all our TPE models (bottom) have much better generalization results with the use of thermal time. TPE-Recurrent shows the best performance by being learnable and capturing the temporal development in GDD, increasing F1 on average by $+15.3\%$ over the default PSE+LTAE~\cite{garnot2020lightweight} model and $+3.4\%$ over the ShiftAug~\cite{nyborg2021timematch} augmented model.
Our TPE greatly improves generalization, but there is still a gap to the upper-bound performance. 
TPE addresses the temporal shifts between regions but does not account for changes in the spectral signature of crops, which can be caused by differences in \eg the topography, soil, or varieties of the cultivated crop type.
We leave this direction to future work.

%other regional changes which could impacts the spectral signature of crops, such as the 

\paragraph{Analysis of results.}
We observe that the default PSE+LTAE with calendar time generalizes worst, obtaining an F1 score of $68.0\%$ on average. Interestingly, simply removing the positional encoding outperforms the baseline significantly, leading to an average performance increase of $+7.9\%$. Since this model variant is given no information about the order of images in the SITS, it is 
also invariant to temporal shifts, which explains the performance increase. 
However, without positional information, the model should not be able to model the class-wise timing differences, which should degrade performance. But the performance increase indicates the model is able to do so.
We argue that this is because the model is able to extract some positional information from the SITS. For example, satellite images taken during the winter differ from those during the summer, enabling the model to extract some degree of temporal order.
However, in the case that two images at different times appear similar, the extracted positions can be ambiguous, which is avoided by providing explicit positional information.
This is also indicated by the result of ShiftAug~\cite{nyborg2021timematch}, where calendar time is augmented with random temporal shifts, which further increases the F1 results by $+11.9\%$ on average over the baseline, outperforming no positional encoding by $+4.0\%$. This indicates that direct positional information is indeed important to the crop classification task to avoid ambiguous order information from images only.
%-- The result of ShiftAug also highlights the importance of accounting for temporal shifts when training with calendar time.
%-- handling the temporal shift when training with calendar time; if not, the model is not robust to temporal shifts.

\begin{table}[t]
\begin{center}
\begin{tabular}{@{}lc@{}}
\toprule 
Method & Training time (s/epoch) \\
\midrule
%PSE+LTAE & 16.1  \\
%No pos &  15.5 \\
%ShiftAug &  16.1 \\
TPE-Sinusoidal &  16.1 \\
TPE-Concat &  \textbf{15.5} \\
TPE-Fourier &  16.4 \\
TPE-Recurrent &  17.2  \\
\bottomrule
\end{tabular}
\end{center}
\caption{The training time of TPE in seconds per training epoch.}
\label{tab:timing_results}
\end{table}

In comparison, our TPE models outperform all calendar time models. 
This highlights the benefits of using thermal time for reducing the temporal shift between different regions without introducing any augmentations, while also providing explicit positional information for modelling the class-wise timing differences.
The TPE-Sinusoidal model is the default PSE+LTAE model but where calendar time positions are replaced with thermal time.
This simple change significantly improves the F1 generalization results by $+14.1\%$ on average.
Learning a sinusoidal PE with TPE-Fourier, however, is not beneficial, resulting in a decrease in F1 compared to TPE-Sinusoidal by $-1.8\%$.
TPE-Concat learns embedding and positional representations jointly in the PSE module, and obtains comparable results to TPE-Sinusoidal, with
higher F1 ($+0.6\%$) but lower OA ($-0.4\%$). But as TPE-Sinusoidal introduces extra computation because of the sinusoidal encoding function, TPE-Concat is computationally more efficient as shown in Table~\ref{tab:timing_results}.
This indicates that the approach of adding positional encodings to input embeddings common in natural language processing may be unnecessary for SITS classification. 
TPE-Recurrent learns a positional encoding that captures the development in GDD, leading to an increase in F1 of $+1.2\%$ over TPE-Sinusoidal. TPE-Recurrent thus shows the best performance but also introduces sequential computation which increases computation requirements as shown in Table~\ref{tab:timing_results}.
We suggest the choice of TPE method is a trade-off between performance and efficiency. Practitioners can easily implement TPE-Concat by concatenating thermal time in PSE~\cite{garnot2020lightweight}, and enjoy improved generalization and efficiency. If more computation can be afforded, TPE-Recurrent offers the best results.

%Next, we analyze the variants with learnable positional encodings. Concatenating the GDD positions with the input embeddings and modulating the embedding by an MLP enables the model to learn a representation different from the fixed vector of sinusoidal PE. The results, however, are comparable, with the concatenated version increasing F1 by $+0.6\%$ on average, but decreasing OA by $-0.4\%$. The generalized version of sinusoidal PE, Fourier PE, which learns the sine/cosine functions decreases the F1 results compared to the standard sinusoidal PE by $-0.8\%$ on average, indicating that the representation of GDD by sines is not beneficial. 
%
      
\subsection{Visual Analysis}
To better understand how TPE obtains improvements, we visualize in Figure~\ref{fig:tsne} t-SNE~\cite{van2008visualizing} embeddings of features output by the LTAE. 
For TPE methods, we observe denser and better separated clusters, indicating better class separation by accounting for temporal shifts.
%Confusion between meadow and unknown can be observed across all methods. This is expected due to the high overlap between these two classes---if a label is not available, the crop type of a parcel is often similar to meadow.
For the baseline PSE+LTAE model~\cite{garnot2020lightweight}, we observe some classes are well clustered despite the temporal shift, such as corn and winter rapeseed, indicating these classes are less impacted by temporal shifts. Others are mixed, such as spring barley/horsebeans and winter wheat/winter triticale.
We observe that temoving the PE results in less dense clusters. Particularly, the clusters for spring barley and winter barley overlaps. This could indicate difficulties in resolving class-wise temporal shifts, since these are better separated with ShiftAug~\cite{nyborg2021timematch}.

\section{Conclusion}
\label{sec:conclusion}
In this work, we propose Thermal Positional Encodings (TPE) to address the temporal shift issue of SITS classifiers and improve generalization. 
While existing work uses calendar time, our TPE uses thermal time, which enables models to account for the varying rates of crop growth in different climates and thereby address the temporal shift issue. We propose different methods to positional encode thermal time, including fixed and learned approaches. 
On a parcel classification dataset with SITS from four different European regions, we demonstrate that TPE significantly improves generalization compared to existing methods.

\paragraph{Acknowledgements.}
\acknowledgements{}
%The work of Joachim Nyborg was funded by the \emph{Innovation Fund Denmark} under reference \emph{8053-00240}.
%We acknowledge the E-OBS dataset from the EU-FP6 project UERRA (\url{https://www.uerra.eu}) and the Copernicus Climate Change Service, and the data providers in the ECA\&D project (\url{https://www.ecad.eu}).

%%%%%%%%% REFERENCES
{\small
\bibliographystyle{ieee_fullname}
\bibliography{egbib}
}

\end{document}